\documentclass[]{article}
\usepackage{comment}
\usepackage{hyperref}
\usepackage{apacite}
\usepackage{graphicx}
\usepackage{amsmath}
\usepackage{amssymb}
\usepackage{verbatim} 
\usepackage{natbib}
\usepackage{bm}
\usepackage{lineno}
\usepackage{hyperref}
\usepackage{placeins}
\usepackage{caption}
\usepackage{mwe}
\usepackage[super]{nth}
\usepackage[ruled,vlined]{algorithm2e}
\usepackage{subcaption}
\usepackage{authblk}
\hypersetup{
  colorlinks=true,
  citecolor=blue,
  linkcolor=blue,
  filecolor=magenta,   
  urlcolor=cyan,
}
\usepackage[left=30mm, right=30mm, top=30mm, bottom=30mm]{geometry}
\title{Generation of non-stationary stochastic fields using Generative Adversarial Networks}

\author[1]{Alhasan Abdellatif \footnote{aa519@hw.ac.uk}}
\author[1]{ Ahmed H. Elsheikh\footnote{
		a.elsheikh@hw.ac.uk}}

\author[2]{Daniel Busby}

\author[2]{Philippe Berthet}

\affil[1]{Heriot-Watt University}

\affil[2]{TotalEnergies}

\begin{document}
\maketitle
\sloppy

\begin{abstract}
In the context of generating geological facies conditioned on observed data, samples corresponding to all possible conditions are not generally available in the training set and hence the generation of these realizations depends primary on the generalization capability of the trained generative model. The problem becomes more complex when applied on non-stationary fields. In this work, we investigate the problem of using Generative Adversarial Networks (GANs) models to generate non-stationary geological channelized patterns and examine the models generalization capability at new spatial modes that were never seen in the given training set. The developed training method based on spatial-conditioning allowed for effective learning of the correlation between the spatial conditions (i.e. non-stationary maps) and the realizations implicitly without using additional loss terms or solving optimization problems for every new given data after training. In addition, our models can be trained on 2D and 3D samples. The results on real and artificial datasets show that we were able to generate geologically-plausible realizations beyond the training samples and with a strong correlation with the target maps.
\end{abstract}

\section{Introduction}

Generating stochastic fields has many applications in geosciences and reservoir management. Modelling these fields at the reservoir scale is an essential step for addressing uncertainty quantification or inverse problems in the subsurface. One of the classical approaches is the Multiple Point Statistics (MPS) algorithms~\cite{strebelle2002conditional} which were designed for geo-statistical simulation based on spatial patterns in a training image. Many variants of MPS have been developed over time such as direct sampling techniques~\cite{mariethoz2010direct} and cross-correlation based methods~\cite{tahmasebi2013cross}. The trained non-parametric model can be used to generate realizations constrained to well and seismic data~\cite{hashemi2014channel,rezaee2017integration,arpat2007conditional,tahmasebi2014ms}. However, MPS algorithms suffer from many limitations such as the high computational cost~\cite{li2016patch},~limited variability \cite{emery2014can} and inability to model complex non-stationary patterns~\cite{zhang2019generating}.

Following the success of deep learning in computer vision, recent published work have considered deep generative models such as Generative Adversarial Networks (GANs)~\cite{goodfellow2014generative} for generation of stochastic fields. A main advantage of GANs over Multiple Point Statistics (MPS) is that the trained model is used to parametrize a high-dimensional sample using a low-dimensional input. This allowed GANs to be used for a variety of tasks, for example, GANs were used to model 3D structure of porous media~\cite{mosser2017reconstruction}, the low-dimensionality of GANs latent vector was utilized to perform parametrization of the spatial permeability fields in the subsurface~\cite{chan2017parametrization} and geostatistical inversion was performed after training GANs models on 2D and 3D categorical samples~\cite{laloy2018training}. 

GANs have also been used to generate geological realizations conditioned on hard data (e.g., point measurements at wells) and soft data (e.g., probability maps). Approaches for generation of conditioned stochastic realizations could be classified into two categories: post-GANs and concurrent-GANs. In post-GANs approaches, a new optimization problem is solved after training GANs where the latent vector is searched to find realizations that match the target data. For example, gradient descent method was used in~\cite{dupont2018generating,zhang2019generating}, a Markov Chain Monte Carlo sampling algorithms were used in~\cite{nesvold2019geomodeling,laloy2018training} and~\cite{chan2019parametric} trained an inference network to map the normally distributed outputs to a distribution of latent vectors that satisfies the required conditions. The main drawback of using post-GANs approaches is the additional cost needed to solve the second optimization problem which can often be expensive. In addition, we would need to solve different problems for every new observed data (e.g. new condition). 

In concurrent-GANs approaches, the training of GANs is modified to pass the conditional data to the GANs generator network. After training, the trained generator can then simulate realizations based on the input data without the need to solve another optimization problem. In~\cite{ABDELLATIF2022105085}, conditional GANs were used to generate unrepresented global proportions of geological facies. Cycle-consistent GANs~\cite{zhu2017unpaired} has been used for domain mapping, for example mapping between physical parameters and model state variables~\cite{sun2018discovering} and mapping between seismic data and geological model~\cite{mosser2018rapid}. In \cite{zhong2019predicting}, a GANs model with a U-net architecture~\cite{ronneberger2015u} was used to map high-dimensional input to $\textrm{CO}_{2}$ saturation maps. However, the one-to-one mapping using Cycle-GANs or U-net architecture is not suitable for generating multiple stochastic facies conditioned on a single observed data.

In~\cite{song2021gansim}, condition-based loss functions were used to condition facies on hard data and global features and later they extended the method for spatial probability maps in~\cite{song2021bridging}. Pix2pix method~\cite{isola2017image} has been used in~\cite{pan2021stochastic} for geophysical conditioning by adding additional losses for seismic and well log conditions. However, condition-based losses require designing manual functions that compute the consistency between the generated samples and target conditions (e.g., computing facies frequency for the generated realizations to mimic real probability maps~\cite{song2021bridging}). The design of such function is arbitrary and this is conceptually different than GANs, where the learning is done implicitly from the training data by joint training of both the generator and the discriminator which tells what is good versus bad samples. Moreover, including additional losses in GANs relies on careful weighting between the conditions losses and the original loss in GANs which requires extensive hyper-parameters search (e.g., see Fig 6 in~\cite{song2021gansim}).

In this work, we developed a concurrent-GANs approach that conditions the geological realizations on spatial maps, describing the distribution of facies proportions across the spatial domain, without using condition-based losses. Further, we were able to generate new realizations that match spatial maps never seen in the training samples; such generalization capability is very useful especially when the characteristics of actual reservoirs differ significantly from those of the available training samples. To model the spatial modes, we used conditional GANs (cGANs)~\cite{mirza2014conditional} where each spatial map is provided as a condition to the neural networks. By letting the maps modulate the generator layers in a spatial setting through the SPADE algorithm~\cite{park2019semantic}, the GANs models learned the correlation between the spatial conditions and the samples implicitly. The developed training approach solves a single optimization problem (i.e., a concurrent-GANs method) without using condition-based losses and hence there is no need to design condition-consistency functions or to perform careful hyper-parameter search for balancing weights. Results on 2D and 3D samples show a strong correlation between the generated samples and the target maps as well as the trained models were able to generalize to unseen spatial modes. 

The rest of the paper is organized as follows: in section~\ref{method}, we discuss the algorithm of conditional GANs used in our experiments and we present the training datasets and the implementation details. In section~\ref{results}, the results of the experiments are shown. Finally, conclusions are provided in section~\ref{conlcusion}.

\section{Method and Materials}
\label{method}
\subsection{Method}

Generative adversarial networks (GANs)~\cite{goodfellow2014generative} are trained to learn the underlying distribution of training samples. It consists of two convolutional neural networks: a generator $G$ and a discriminator $D$. The generator maps a random noise $z$ to a realization $G(z)$ while the discriminator takes samples from the real and the generated sets and is optimized to output the probability of the samples being real (i.e., not generated by the generator). The generator is then optimized such that the generated samples have high probability $D(G(z))$. The two networks are trained in an adversarial setting defined by the objective function $V(G,D)$:

\begin{equation}
	\label{adv_eq}
	\underset{G}{\min} \, \underset{D}{\max} \, V(G,D) 
	=\mathbb{E}_{x\sim p_x} [\log D(x)] + \mathbb{E}_{z\sim p_z} [\log (1-D(G(z)))].
\end{equation}

Given a spatial map $\textbf{M}$ that describes the spatial distribution of a geological facies (e.g., channels), we can direct the generated samples to match a particular spatial map $\textbf{M}$ by using a conditional GANs~\cite{mirza2014conditional} method, where the condition $\textbf{M}$ is passed to both the generator network $G$ and discriminator network $D$ during training. Similar to the concurrent-GANs methods, after training we can generate multiple realizations conditioned on $\textbf{M}$ by simply passing $\textbf{M}$ and different latent $z$ vectors to the generator without solving a new optimization problem. The discriminator would then output the conditional probability of the sample being real given its input map. The objective function of conditional GANs is then:
\begin{equation}
	\underset{G}{min} \, \underset{D}{max} \, V(G,D) =\mathbb{E}_{x\sim p_x} [\log D(x|\textbf{M})]
	+ \mathbb{E}_{z\sim p_z} [\log (1-D(G(z,\textbf{M})|\textbf{M}))].
\end{equation}

To accommodate for the spatial nature of the map, we followed the spatially adaptive de-normalization (SPADE) conditioning method developed in \cite{park2019semantic}, where a segmentation mask is used to modulate the generator layers such that it could generate natural images based on the mask.

In our work, we replaced the categorical mask by a continuous map representing the spatial proportions of the channels, the maps were calculated for each sample before training. The SPADE method goes as following, for each layer $i$ and channel $c$ in the generator, each activation $h_{i,c,x,y}$ ($h_{i,c,x,y,z}$ in case of 3D samples) is normalized using the mean $\mu_{i,c}$ and standard variation $\sigma_{i,c}$ computed over both batch instances and channel spatial locations. The result is then de-normalized spatially, i.e., per spatial position, using parameters $\gamma$ and $\beta$ which are learnable functions of $\textbf{M}$. The calculation of SPADE algorithm for the 2D case is shown in following equation:
\begin{equation}
	\label{spade_eq}
	\hat{h}_{i,c,x,y}(\textbf{M}) = \gamma_{i,c,x,y}(\textbf{M})\frac{h_{i,c,x,y}-\mu_{i,c}}{\sigma_{i,c}} +\beta_{i,c,x,y}(\textbf{M}), \, \, \textbf{M} \in \mathbb{R}^{H\times W},
\end{equation}

where the learnable parameters $\gamma$ and $\beta$ were obtained using two successive convolutional layers, separated by a ReLU activation function, directly applied to the map $\textbf{M}$ and $H \times W$ is the grid dimension at which the channels proportions were calculated.

We can then modify the spatial proportions of the generated facies by modifying $\textbf{M}$ which will in turn modulate the generator activations through $\gamma$ and $\beta$. Since each layer of the generator operates at different resolution, the map $\textbf{M}$ is down-sampled (or up-sampled in case $\textbf{M}$ has a lower resolution) to match the resolution of the features at each layer. When extending the problem of 3D samples, the proportions maps will be calculated for all 3 dimensions. Equation \ref{spade_eq} for conditioning on 3D $\textbf{M}$ becomes:

\begin{equation}
	\label{spade_eq_3d}
	\hat{h}_{i,c,x,y,z}(\textbf{M}) = \gamma_{i,c,x,y,z}(\textbf{M})\frac{h_{i,c,x,y,z}-\mu_{i,c}}{\sigma_{i,c}} +\beta_{i,c,x,y,z}(\textbf{M}), \, \, \textbf{M} \in \mathbb{R}^{H\times W \times D}.
\end{equation}

In the discriminator, features computed from the map $\textbf{M}$, using convolutional layers, are concatenated with spatial features computed from the input image at an intermediate layer of the discriminator. The intermediate layer is chosen such that its resolution matches the resolution of $\textbf{M}$. For example, if the input images are of resolution $64\times64$ and $\textbf{M}$ is $4 \times 4$ then the concatenation is done at the fourth layer.
\subsection{Implementation Details}
\label{Imp}
All models are based on the ResNet architecture~\cite{he2016deep} following~\cite{gulrajani2017improved}. The discriminator halves the features resolution and doubles the number of features channels after each layer before it outputs a single value indicating the probability of the image being real. We apply spectral normalization to the discriminator's weights~\cite{miyato2018spectral} and the self-attention mechanism~\cite{zhang2019self} in both the generator and discriminator at an intermediate layer of resolution 32$\times$32. For all experiments, the models are trained using the Adam optimizer with fixed learning rate of $0.0002$ for both networks and a batch-size of 32. The latent vector $z$ is sampled from a multivariate standard normal distribution of dimension $128$. The final checkpoint used is based on an exponential moving average of the generator weights with a decaying factor of 0.999 following~\cite{brock2018large}. When updating the generator we used $-\mathbb{E}_{z\sim p_z} [\log (D(G(z)))]$ as proposed in~~\cite{goodfellow2014generative}.

\begin{figure*}[h]
	\centering
	\includegraphics[width=\linewidth]{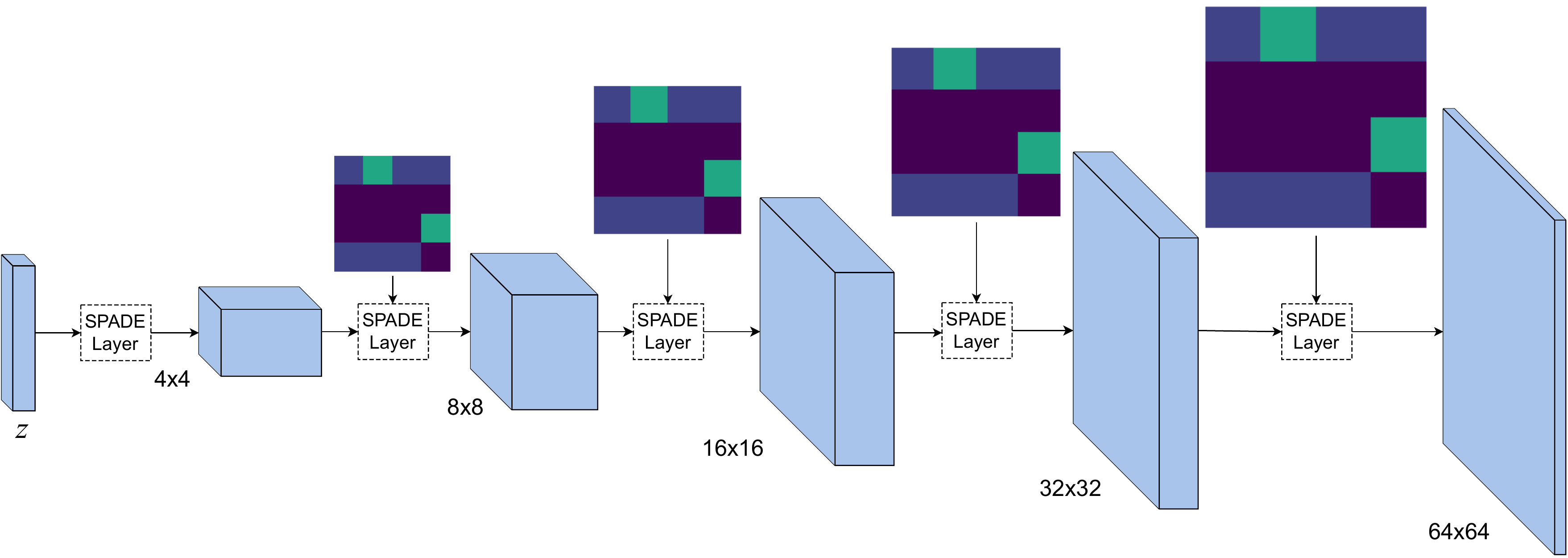}
	\caption{Generator architecture: the stochastic input $z$ is projected into a generated image and the conditioning map is passed to each layer in the generator for spatial modulation. SPADE layer is similar to the one explained in \protect\citeA{park2019semantic}.}
	\label{fig:G_arch}
\end{figure*}

\begin{figure*}[h]
	\centering
	\includegraphics[width=\linewidth]{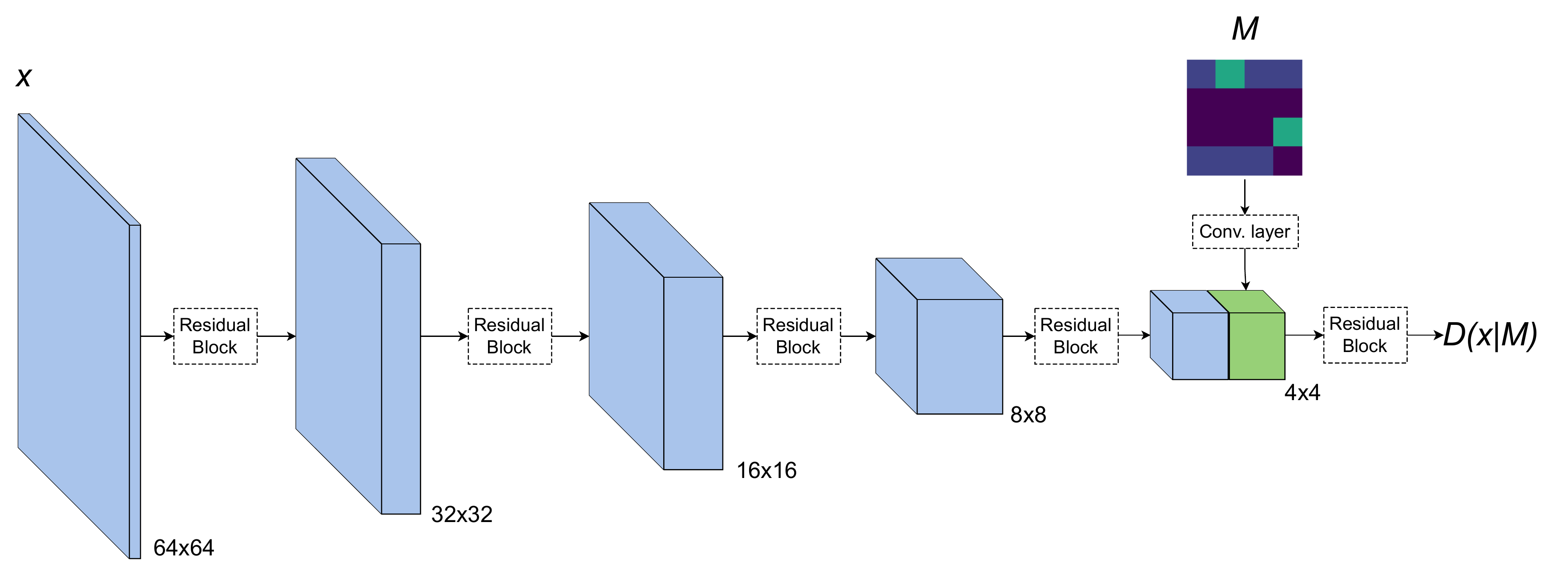}
	\caption{Discriminator architecture: the conditioning map is passed to convolutional layers and the resulting features are concatenated with the input image features (i.e., blue and green features). The discriminator output is the conditional probability of $x$ being real given its corresponding map $\textbf{M}$.}
	\label{fig:D_arch}
\end{figure*}

For the 3D model, 3D convolutions, 3D batch-normalization and 3D up-sampling operations are adapted from the PyTorch framework. In this case, the starting vector $z$ is reshaped to $4\times4\times2$ and passed to the generator to form $64\times64\times32$ 3D images. To handle the size of the 3D samples, we trained our models on 4 parallel GPUs of GeForce RTX 3090. The architectures of the generator and discriminator networks are depicted in Figures \ref{fig:G_arch} and \ref{fig:D_arch}, respectively, for simplicity, we removed the self-attention blocks and showed only the 2D models. 

\subsection{Datasets}
\begin{figure*}
	\centering
	\includegraphics[width=\linewidth]{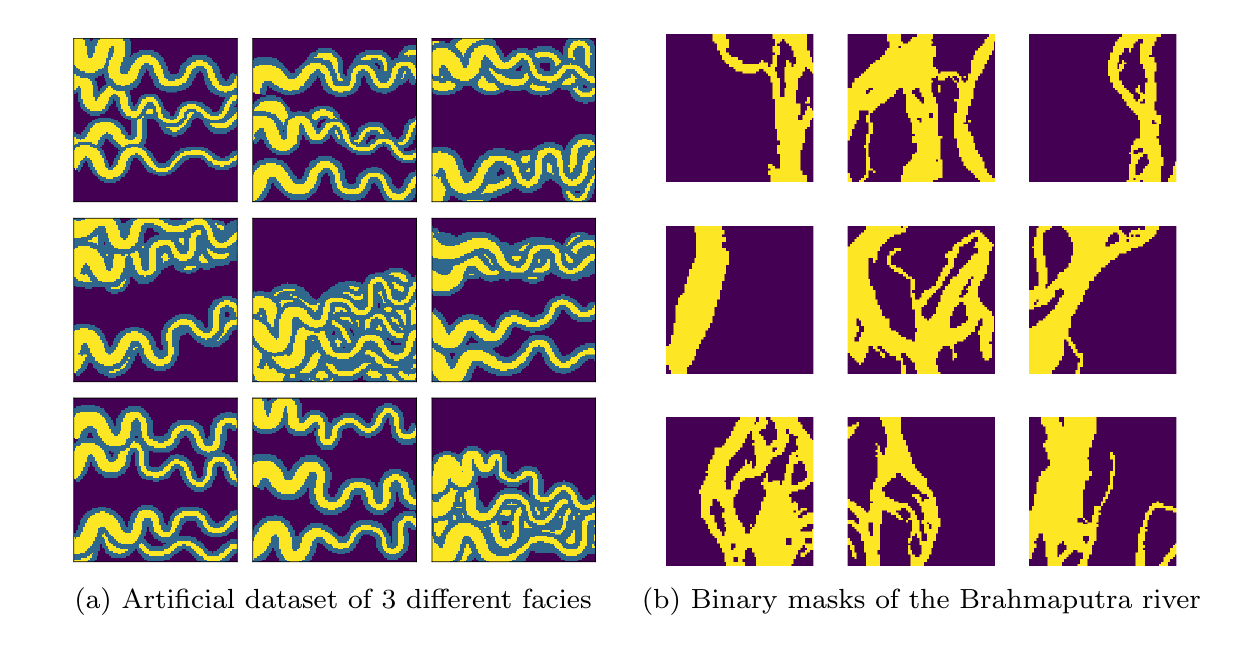}
	\caption{Samples from the 2D datasets used to train GANs models. All images used to train the 2D models are of size $64\times 64$.}
	\label{fig:datasets}
\end{figure*}

In our experiments, we used three datasets to test our models: a) a 2D artificial dataset of 3 facies: channels, levees and background b) a 2D dataset of masks of the Brahmaputra river with binary facies: channels and background and c) a 3D artificial dataset of binary facies: channels and background. Samples from the 2D datasets are shown in Figure \ref{fig:datasets}. The non-stationarity in the datasets are due to the variations in the channels proportions across the spatial domain, we describe the 2D datasets and the preprocessing steps below. Details about the 3D dataset are available in \ref{3D_res}. 

Samples of the first dataset (a) are generated using a geo-modelling tool that mimics depositional environment formation based on random walks~\cite{massonnat2019random}. Horizontal and vertical flipping were performed to increase the sample size from $2,000$ to $8,000$, this added an additional mode in the training set (i.e., large channel proportion on the right side). The Brahmaputra river mask, dataset (b), was obtained from the supplementary data in~\cite{schwenk2020determining}. The large mask of size 13091$\times$11680 was cropped to images of size 256$\times$256 with a stride of 64 then the cropped images were rotated such that they have vertical alignment with the central line of the large mask. The central line was computed using RivGraph library~\cite{schwenk2020determining}. Horizontal and vertical flipping were performed to increase the variations in the training set (increasing sample size from 1788 to 7152). All images of the 2D two datasets were resized to 64$\times$64 resolution to match the networks input.

For each sample in the training set, its the channel proportions map was calculated at a resolution of $\textbf{M} \in 4\times$4 for the 2D samples and $\textbf{M} \in 4\times 4 \times 2$ for the 3D samples. Although they could be calculated at higher resolutions, we chose to mimic the low resolutions usually obtained from seismic surveys, see Figures \ref{fig:AR_res} and \ref{fig:BR_res}. After training, $\textbf{M}$ can be arbitrary selected to mimic the non-stationarity (e.g., $p>=0.4$ is a high proportion and $p<=0.16$ is a low proportion).

\section{Results and Discussion}
\label{results}
\begin{figure*}
	\centering
	\includegraphics[width=\linewidth]{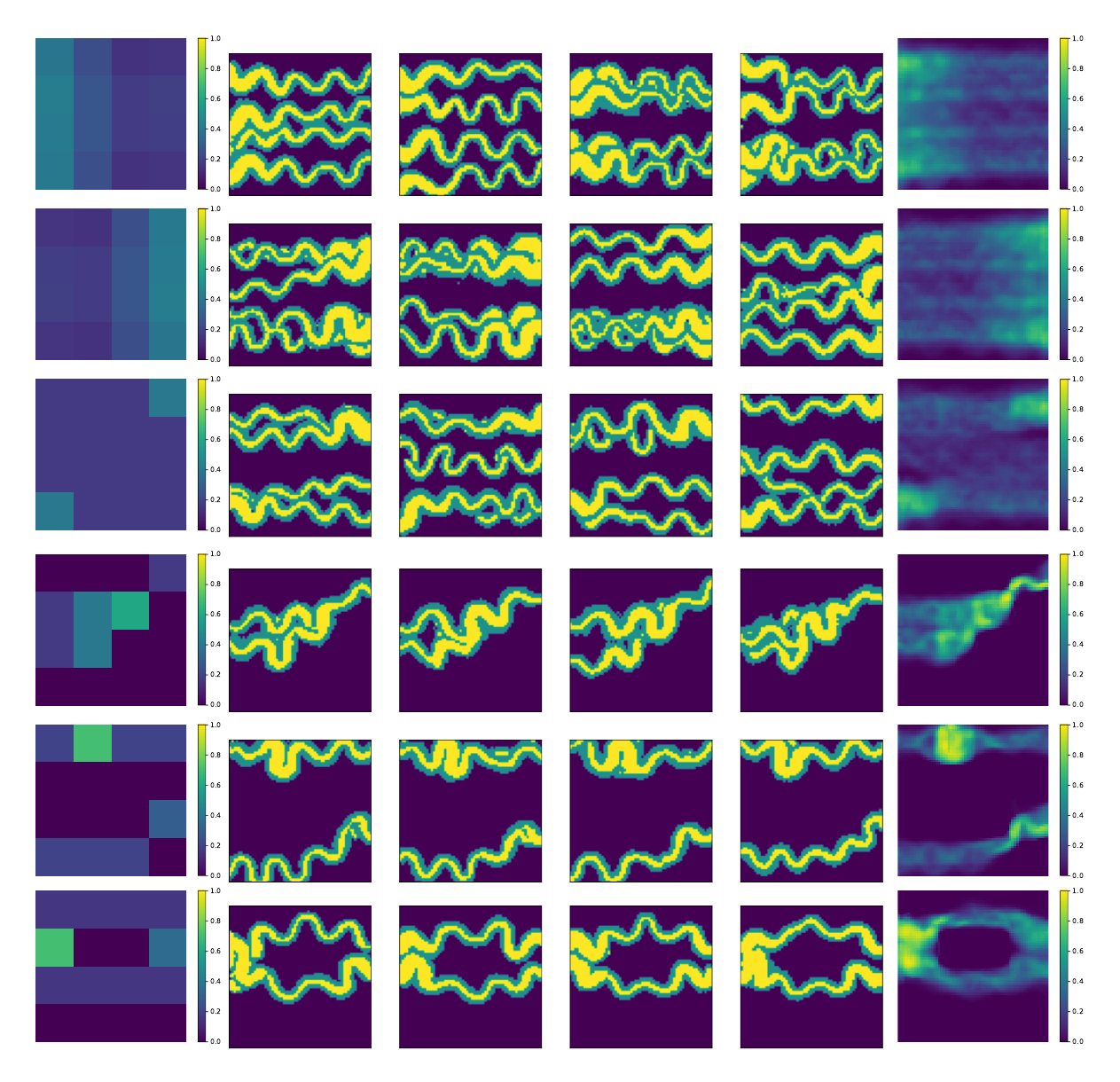}
	\caption{Generated non-stationary realizations on the artificial dataset: the input conditioning maps are in the leftmost columns, the middle columns are the generated samples and the per-pixel mean maps are in the rightmost columns. The last four rows shows generated samples with never seen maps.}
	\label{fig:AR_res}
\end{figure*}

\begin{figure*}
	\centering
	\includegraphics[width=\linewidth]{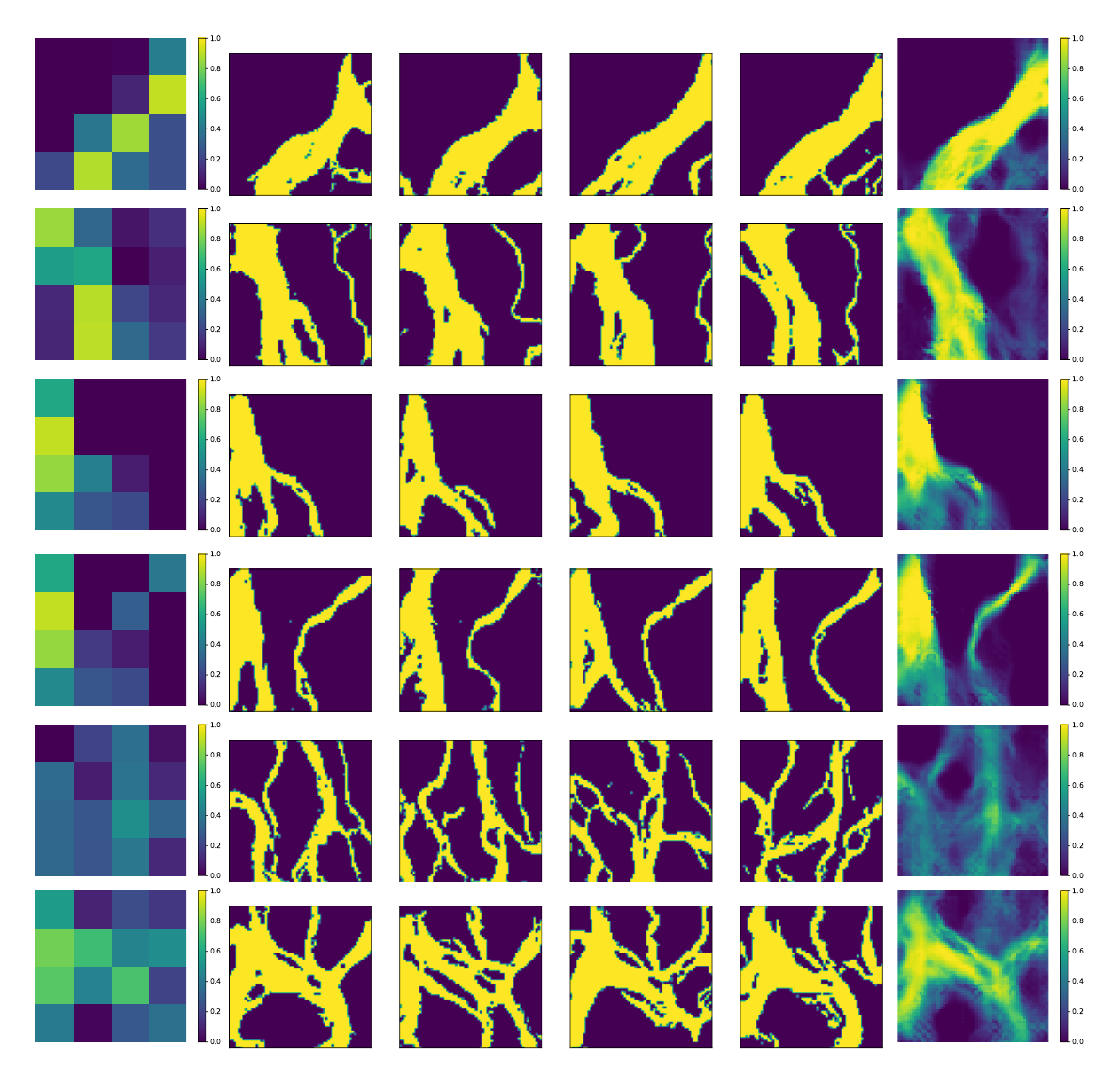}
	\caption{Generated non-stationary realizations on the real masks of the Brahmaputra river: the input conditioning maps are in the leftmost columns, the middle columns are the generated samples and the per-pixel mean maps are in the rightmost columns. The last three rows shows generated samples with never seen maps}
	\label{fig:BR_res}
\end{figure*}

Results on the 2D artificial dataset and the Brahmaputra river masks are shown in Figures \ref{fig:AR_res} and \ref{fig:BR_res}, respectively. The leftmost column shows the conditioning maps $\textbf{M}$, the middle columns are the corresponding generated images $G(z,\textbf{M})$ and rightmost column shows the mean per pixel maps calculated over 2,000 generated samples. For each row, we used the same conditioning map and different $z$ vectors. As shown, the generated realizations exhibit variability due to the randomness of $z$ but overall respect the conditioning map, we show the stochastic variations due to $z$ in \ref{sev_diff}. The results of the 3D models are shown in \ref{3D_res}.

From the results, it is clear that the models were able to generalize successfully to modes never seen in the training datasets. Visually, the generated samples at all modes are geologically-plausible; for example, the channels connectivity is maintained for both datasets as well as the models were able to generate levees surrounding the channels, in the case of the first dataset, irrespective of the location of the channels which indicates that the models have not memorized the training samples.

\begin{figure*}
	\centering
	\includegraphics[width=0.9\linewidth]{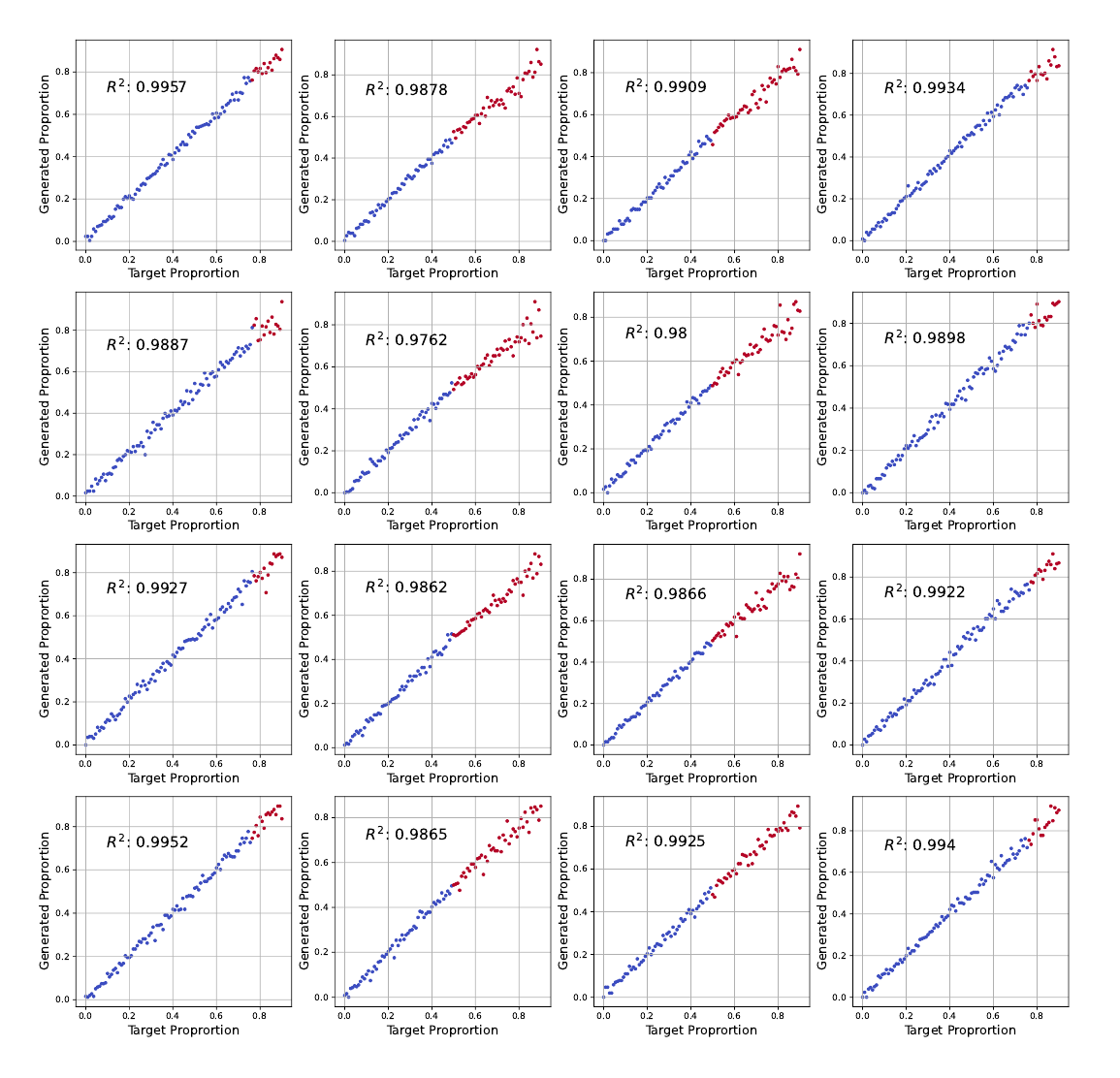}
	\caption{Cross plots between the generated channel proportions and target proportions in the $4\times4$ conditioning maps for the artificial dataset. In each plot, we change the corresponding value in the map and fix all the other values. The blue dots lie within the represented range in the training set while the red ones lie beyond the seen range.}
	\label{fig:dataset_a_cross_plot}
\end{figure*}

\begin{figure*}
	\centering
	\includegraphics[width=0.9\linewidth]{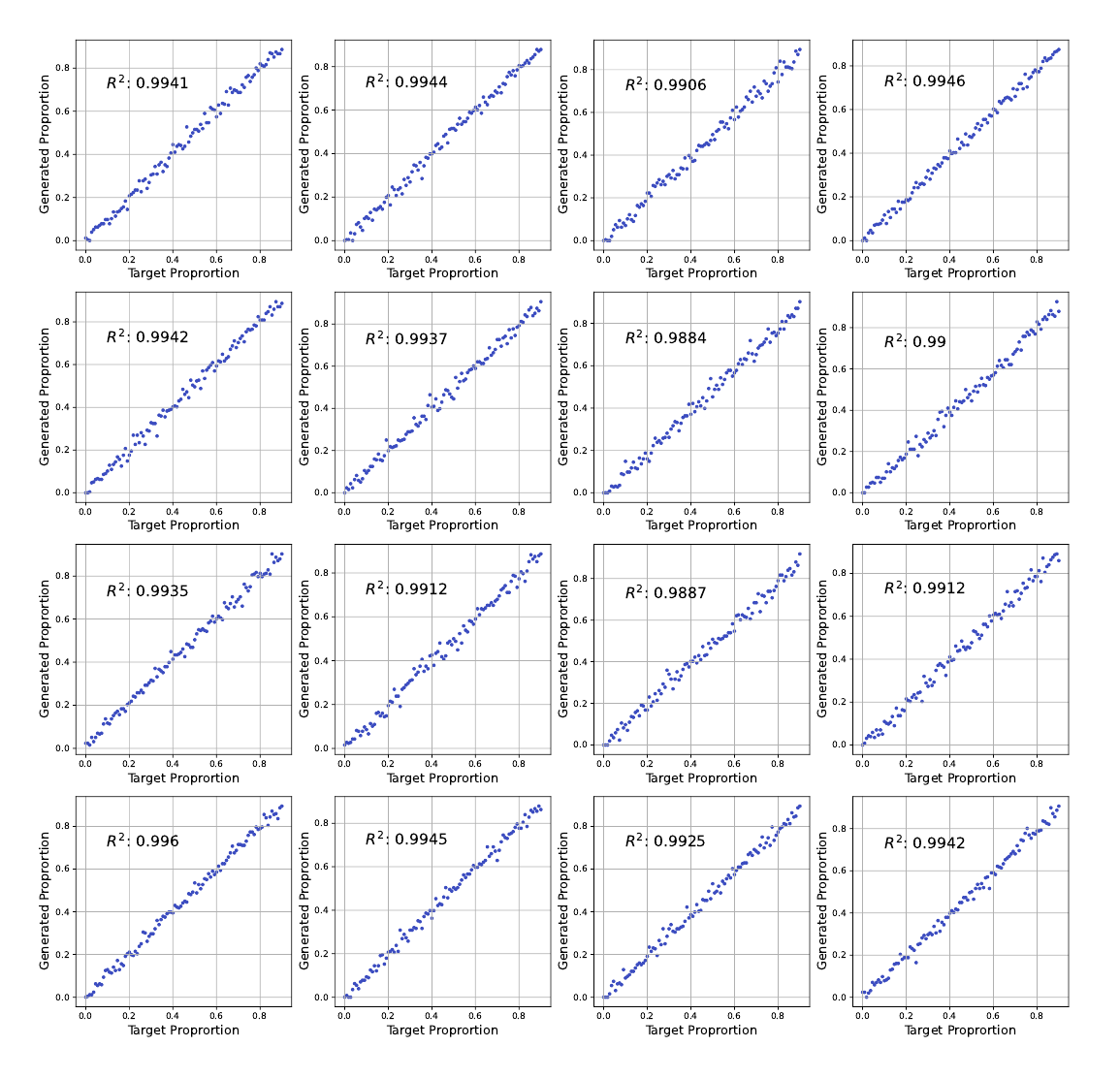}
	\caption{Cross plots between the generated channel proportions and target proportions in the $4\times4$ conditioning maps for the real masks dataset. In each plot, we change the corresponding value in the map and fix all the other values. Notice that all points are blue as the training proportions cover the whole range $[0,1]$.}
	\label{fig:dataset_b_cross_plot}
\end{figure*}

To quantify the correlation between the target proportions maps and the corresponding generated samples, Figures \ref{fig:dataset_a_cross_plot} and \ref{fig:dataset_b_cross_plot} show $4\times 4$ cross plots where each plot represents proportions of each section in the $4\times 4$ grid of $\textbf{M}$. The results show a strong correlation between the target and the generated proportions with an $R^2$ value of almost 1. In addition, the models were able to extrapolate to unseen range of proportions (the red dots in Figure \ref{fig:dataset_a_cross_plot}) in the artificial training set while the proportions for the entire range $[0,1]$ were represented in the real dataset. The generalization capacity of GANs can then be understood from two perspectives:
\begin{enumerate}
	\item The model can generalize to unseen proportions within each section in the grid as shown in Figure \ref{fig:dataset_a_cross_plot}.
	\item The model can generalize to unseen non-stationary patterns over the whole image (combining structures from patches not in the training set) as shown in Figures \ref{fig:AR_res} and \ref{fig:BR_res}.
\end{enumerate}

\begin{figure*}[h]
	\centering
	\includegraphics[width=0.9\linewidth]{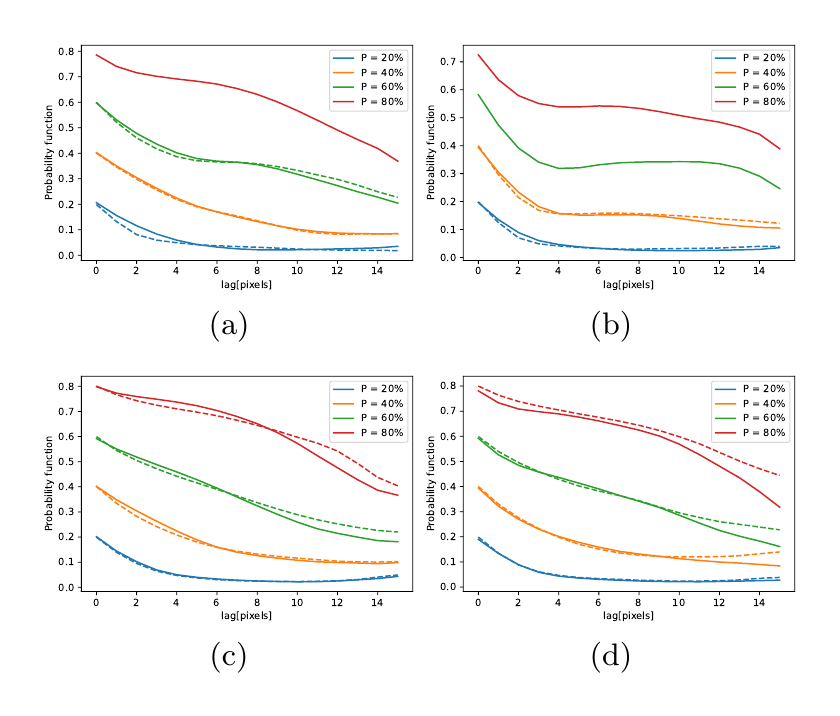}
	\caption{Two-point probability functions calculations for the 2D datasets. The dashed lines indicate calculations from the training samples at the same proportions while the solid lines are for the generated samples. The computation was done for two sections in the $4 \times 4$ grid for the 2D datasets. Figures (a) and (b) show the probability functions calculated for the artificial dataset while Figures (c) and (d) show the probability functions for the real binary masks.}
	\label{fig:pf}
\end{figure*}

We also calculated the two-point probability function which measures the probability that two points, separated by a given distance, have the same channel facies. In Figure \ref{fig:pf}, the function is computed for only two sections (the top left section and the section at the second row and the second column) in the $4\times 4$ grid for the 2D datasets at 4 different conditions. Figures (a) and (b) show the probability functions for the artificial dataset and Figures (c) and (d) are for the real binary masks dataset. As shown, the function for the generated samples (i.e., the solid lines) is consistent with the one for the training samples (i.e., dashed lines). The generalization capability of the model is demonstrated at some unrepresented conditions (e.g., $80\%$ in Figure (a) and $60\%$ and $80\%$ in Figure (d)) where the functions follow the general trend. At large distance, the functions mismatch increases, this could be due to the fact that the model tries to adjust the geological consistency at the boundaries which might be different from the training samples.

\begin{figure*}
	\centering
	\includegraphics[width=0.9\linewidth]{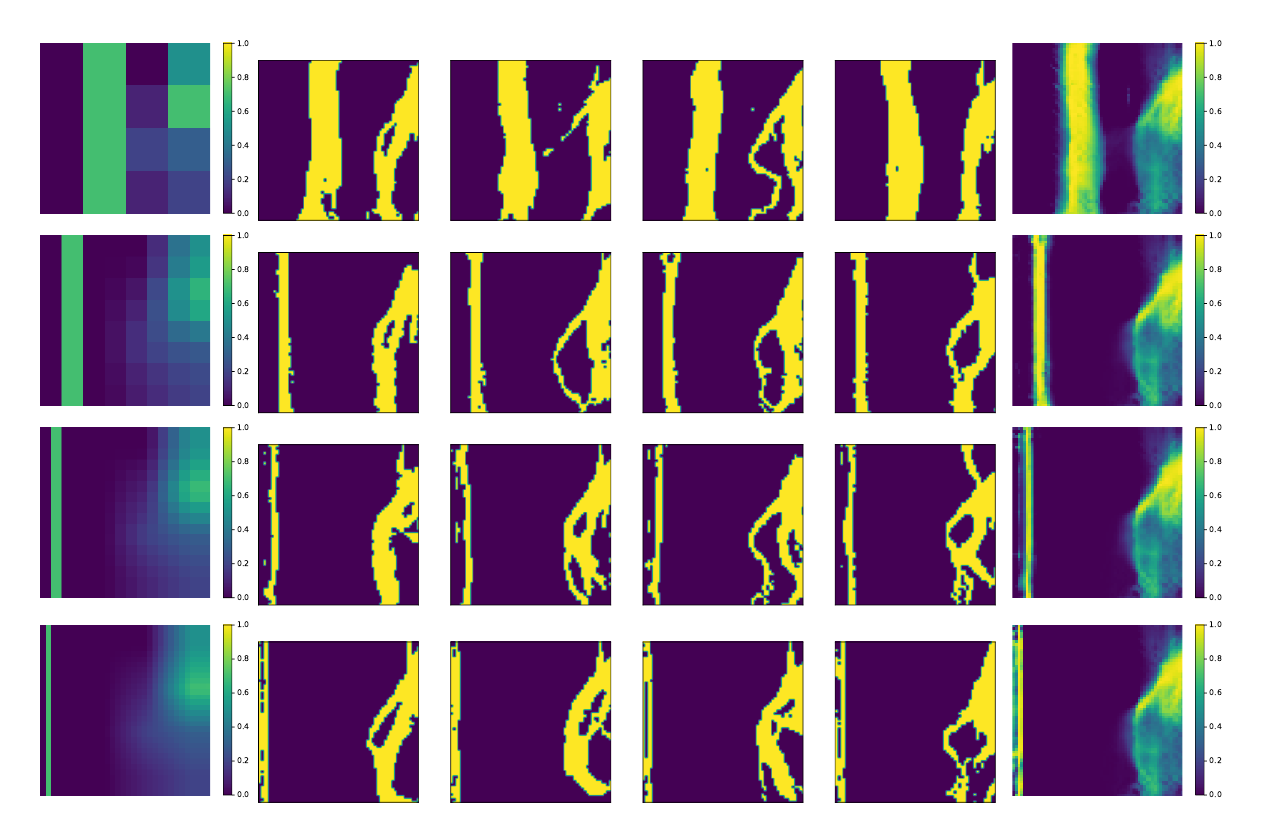}
	\caption{Conditioning realization to high resolution maps. Maps of resolutions $4\times4$, $8\times8$, $16\times16$ and $32 \times 32$ are used in the first, second, third and forth rows, respectively. In all examples, different $z$ vectors are sampled.}
	\label{fig:BR_res_high}
\end{figure*}

Due to the convolutional nature of the conditioning method, after training we can use a conditioning map of a higher resolution than the one used during training. In Figure \ref{fig:BR_res_high}, we show realizations generated using conditioning maps with resolutions of $4\times4$, $8\times8$, $16\times16$ and $32 \times 32$. As shown, despite the fact that the model has been trained only on $4 \times 4$ maps, it is able to generate plausible realizations using maps of higher resolutions. 

To further assess the trained models, we have performed a uniform flow simulation on the artificial samples and the corresponding generated samples and presented the flow statistics in \ref{sev_flow}. 

\section{Conclusion}
\label{conlcusion}
\indent In this work, we investigated the generalization ability of GANs models to generate non-stationary 2D and 3D realizations. The conditioning algorithm used allowed the model to learn the spatial correlations between the target maps and the generated realizations without solving optimization problem for every new observed data and without using arbitrary loss functions. Our models were able to generate consistent 2D and 3D realizations, even with spatial maps never seen during training, on cases representing both artificial and real geological samples. Future work might include generating non-stationary data from stationary training sets and extending the generated field of view to infinite dimensions.

\section{Data Availability Statement}
The artificial datasets used to train the models were generated using a geostastical simulation program based on random walks~\cite{massonnat2019random} found on \url{https://doi.org/10.1016/j.cageo.2022.105085}. The mask of the Brahmaputra river was obtained from the supplementary data of~\cite{schwenk2020determining} found on \url{https://doi.org/10.5194/esurf-8-87-2020} and were pre-processed by the RivGraph library \url{https://github.com/jonschwenk/RivGraph}. The raw 3D dataset can be found on \url{https://github.com/GeoDataScienceUQ/GANRiverI}. The code used to train the GANs models is available on the public repository \url{https://github.com/Alhasan-Abdellatif/NonstationaryGANs}.

\section{Acknowledgment}
\noindent The first author thanks TotalEnergies for the financial support. The authors acknowledge TotalEnergies for authorizing the publication of this paper and would like to acknowledge the use of the flow simulation codes developed by Dr. Shing Chan (Oxford Big Data Institute) during his PhD work at Heriot-Watt university.

\bibliographystyle{apacite}
\bibliography{ref}

\appendix
\section{Generating non-stationary 3D realizations}
\label{3D_res}
In this section, we present the 3D datasets and results of the trained models using the methodology in section \ref{method}.
\subsection{3D Dataset}
\begin{figure*}
	\centering
	\includegraphics[width=\linewidth]{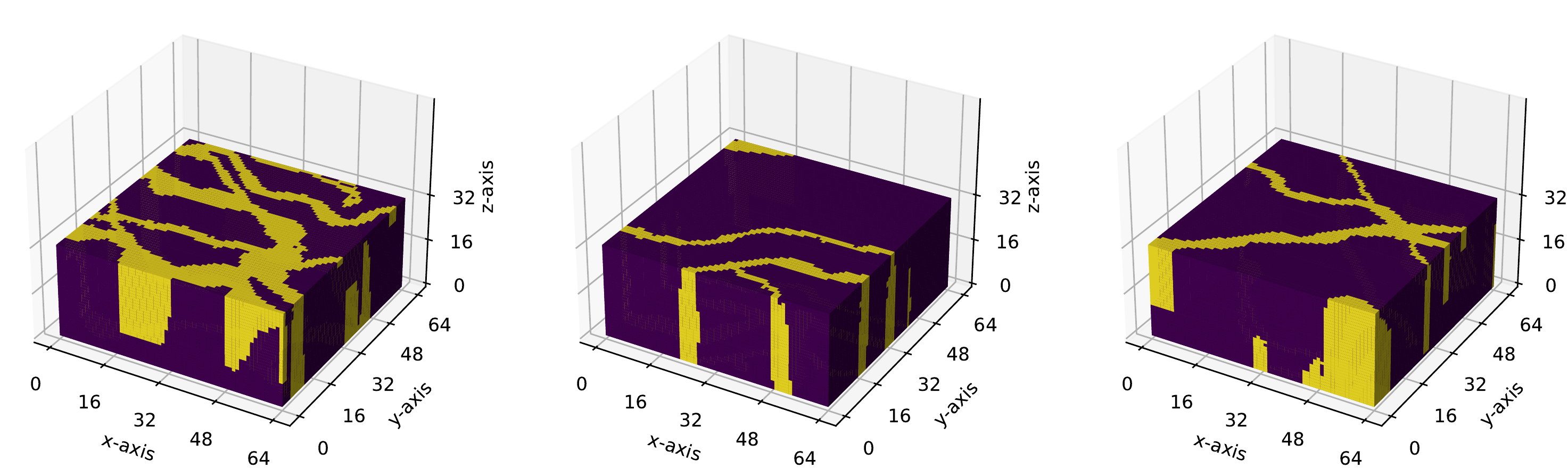}
	\caption{Samples from the 3D training dataset, all images used to train the 3D models are of size $64\times64\times32$.}
	\label{fig:3d_datasets_cubes}
\end{figure*}

\begin{figure*}
	\centering
	\includegraphics[width=\linewidth]{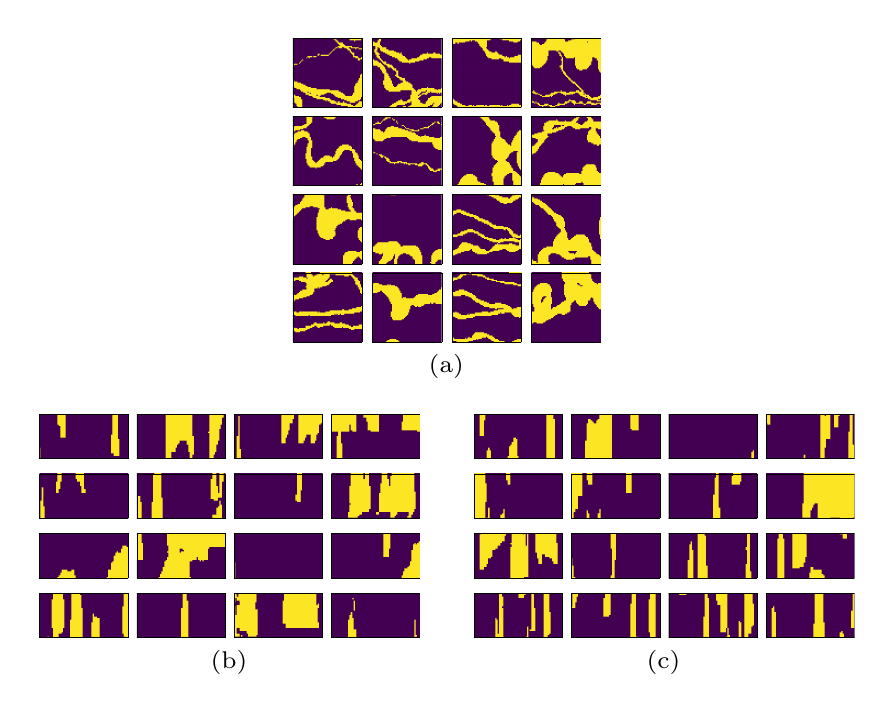}
	\caption{Samples from the 3D training dataset as 2D slices: (a) layers of xy-plane (b) layers of xz-plane (c) layers of yz-plane.}
	\label{fig:3d_datasets}
\end{figure*}

The 3D dataset (c) was obtained from~\cite{sun2023gan} and has been used in~\cite{sun2021comparison} to compare different GANs models. The original dataset is composed of 25 3D images, of size $256\times256\times640$, produced using FLUMY\textsuperscript{TM} computer simulation program and grouped into 5 groups with different avulsion rates. We selected samples from only the first two groups with low avulsion rates and performed cropping on each image such that the cropped images are of size $64\times64\times32$. We used the 3-facies dataset and converted all images to be binary by merging the point bar and channels facies. Finally, flipping was performed to increase the diversity within the dataset. Samples from the dataset are depicted in Figures \ref{fig:3d_datasets_cubes} and \ref{fig:3d_datasets}.
\subsection{Results of the 3D Models}
\begin{figure*}
	\centering
	\includegraphics[width=\linewidth]{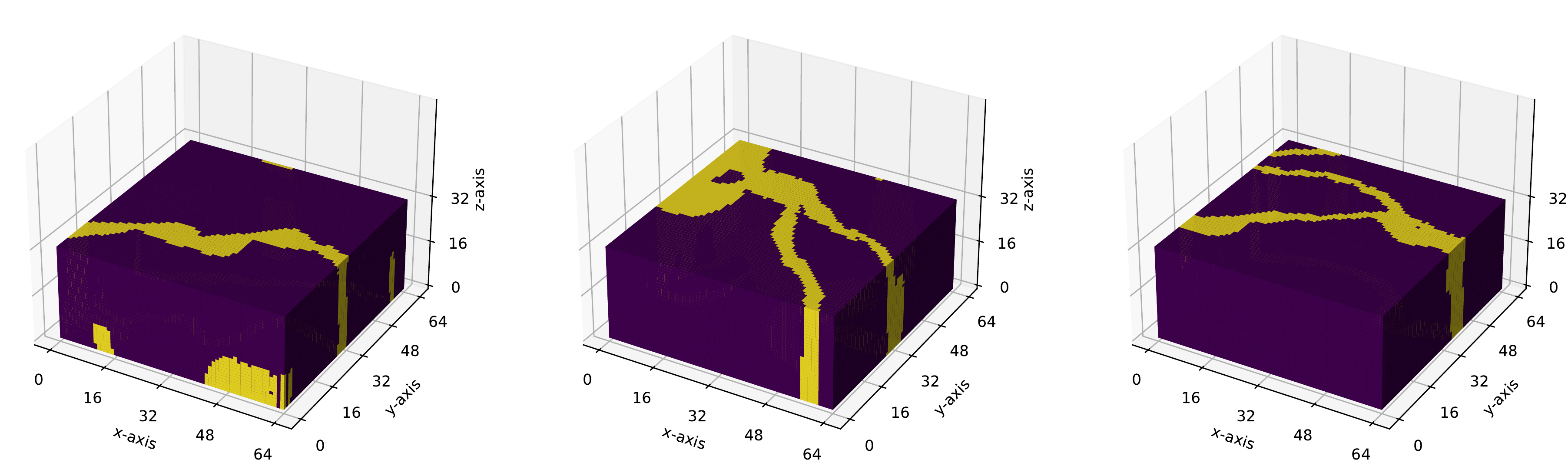}
	\caption{Unconditional generated 3D samples.}
	\label{fig:3d_res_cubes_uncond}
\end{figure*}

\begin{figure*}
	\centering
	\includegraphics[width=\linewidth]{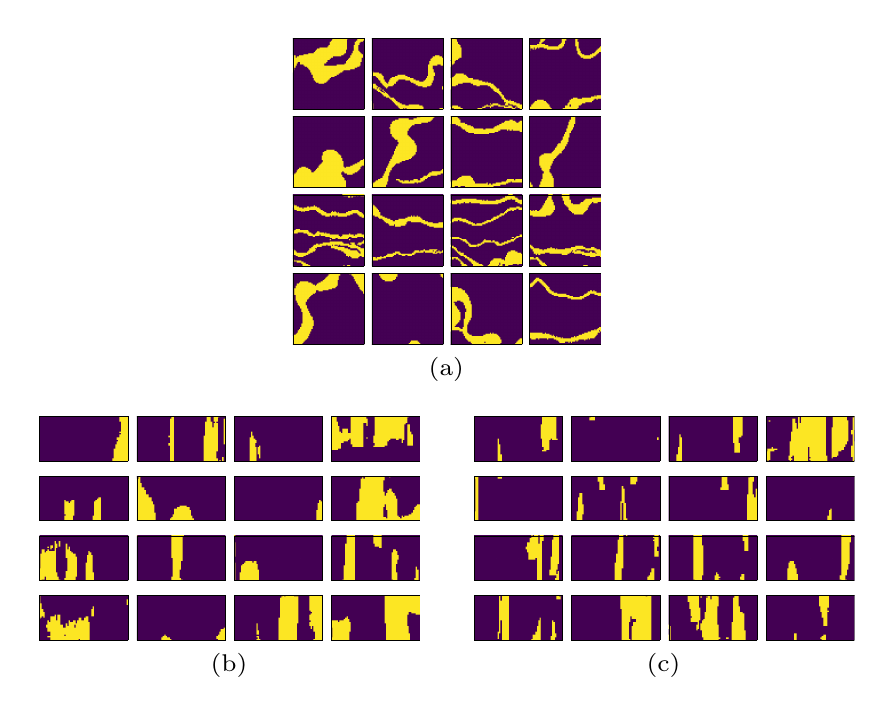}
	\caption{Unconditional generated 3D samples as 2D slices: (a) layers of xy-plane (b) layers of xz-plane (c) layers of yz-plane.}
	\label{fig:3d_res_uncond_slices}
\end{figure*}

\begin{figure*}
	\centering
	\includegraphics[width=\linewidth]{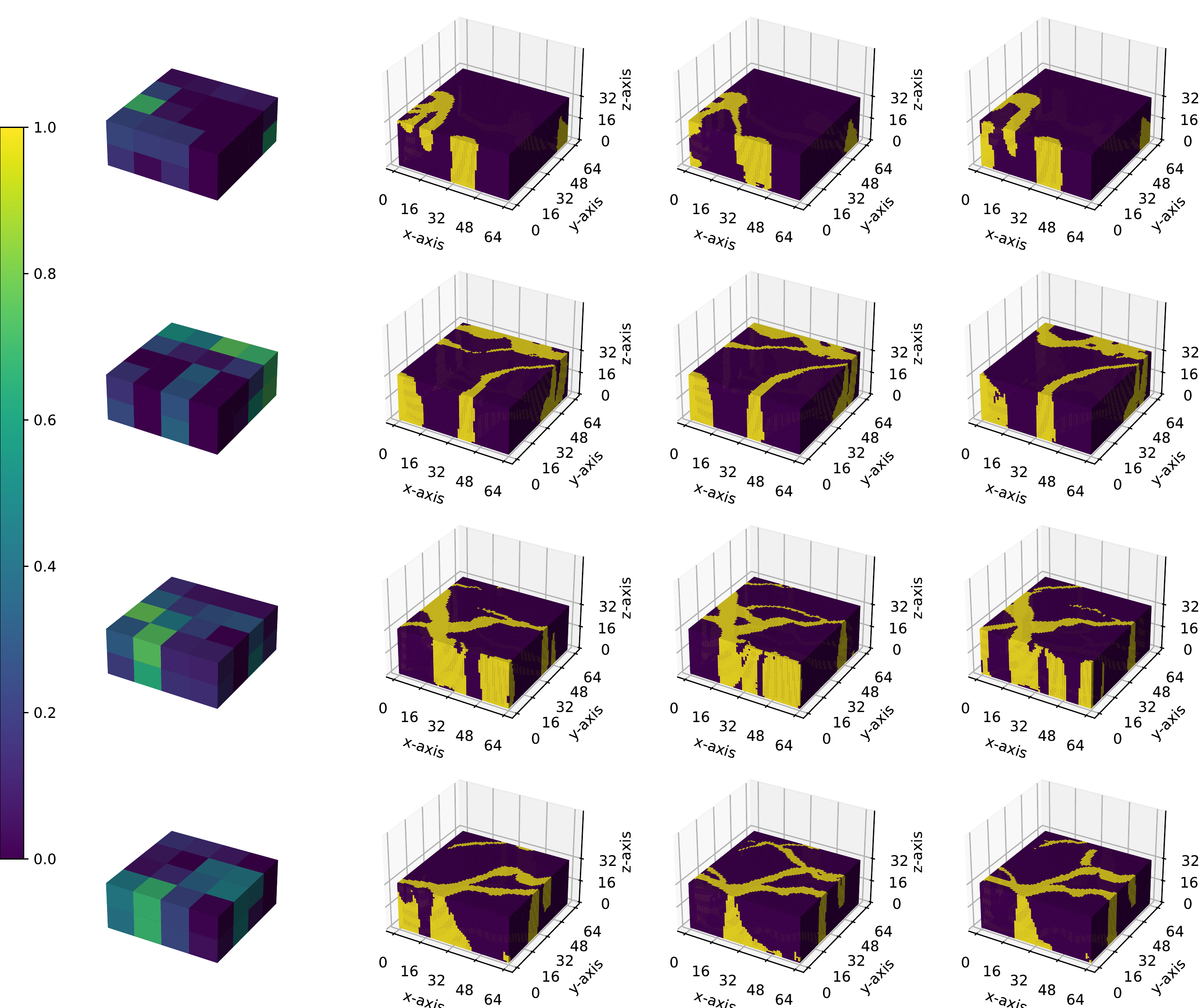}
	\caption{Conditional generated 3D samples. In each row, the first column shows the 3D target map and the remaining columns show the generated 3D realization using different random $z$ and the same $\textbf{M}$.}
	\label{fig:3d_res_cubes_cond}
\end{figure*}

To demonstrate the effectiveness of our models to reconstruct the 3D samples, we trained an unconditional model and show the 3D generated realizations in Figures \ref{fig:3d_res_cubes_uncond} and \ref{fig:3d_res_uncond_slices}. As shown, the images across different planes resemble those found in the training set in Figure \ref{fig:3d_datasets}; which confirms that GANs were able to model the 3D structure of the channels. In Figure \ref{fig:3d_res_cubes_cond}, we present the 3D conditional results: in each row, the first column shows the target conditional map $\textbf{M}$ and the remaining columns show 3 different generated 3D realizations $x_i = G(z_i,\textbf{M})$, where each realization is generated with a different $z_i$ and the same map $\textbf{M}$. From these results, it is clear that our 3D models have learnt a disentangled representation between the $z$ vector which drives the stochastic variation and $\textbf{M}$ which forces the generated samples to respect the given channel distribution.

\begin{figure*}
	\centering
	\includegraphics[width=\linewidth]{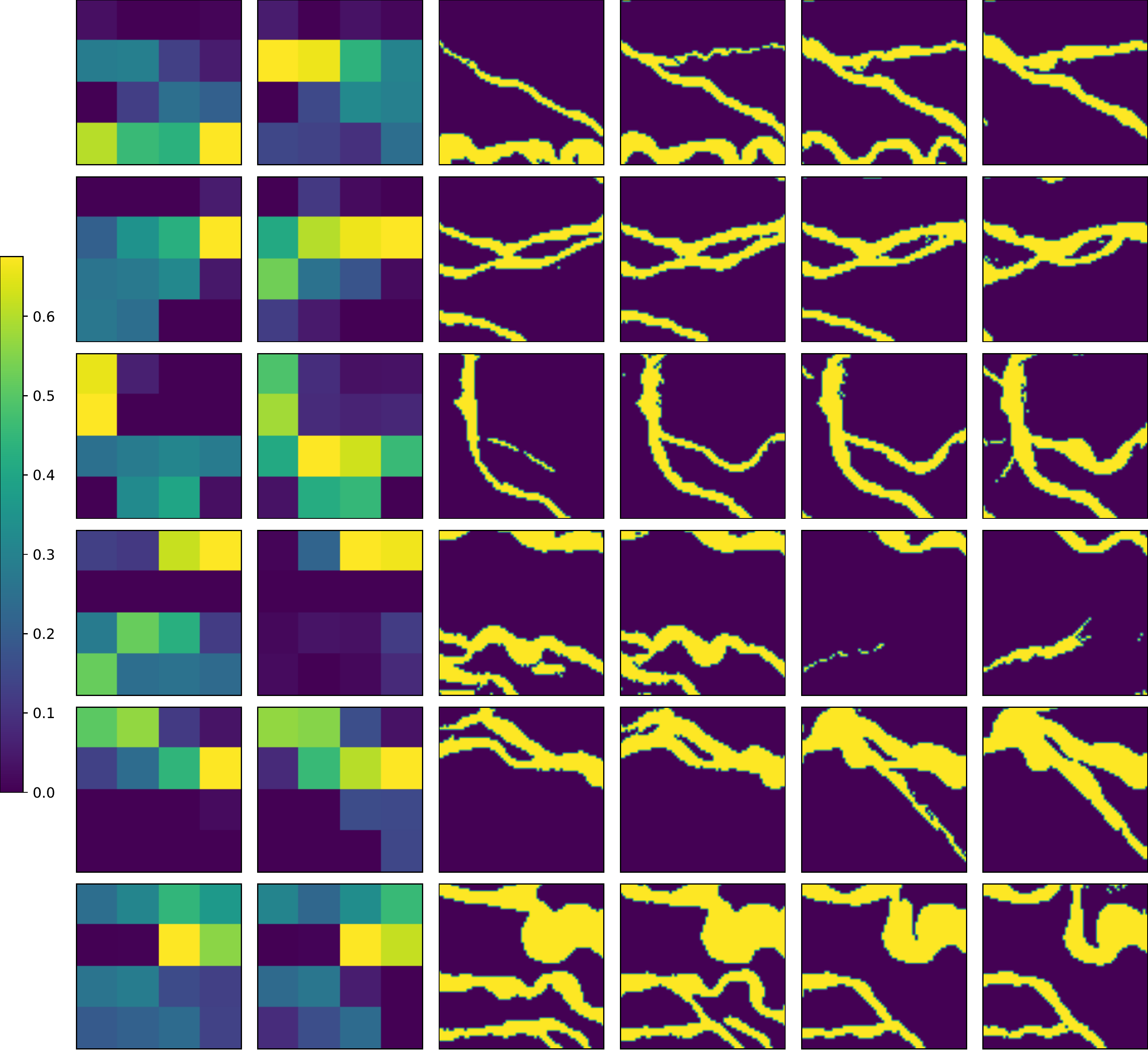}
	\caption{Conditional generated 3D samples with target 3D maps: the first two columns constitute the two layers of the $4\times 4 \times 2$ target maps and the remaining columns show the corresponding generated 3D sample at different layers, namely the \nth{1}, \nth{8}, \nth{16} and \nth{24} layers.}
	\label{fig:con_3d_res_slices}
\end{figure*}

In Figure \ref{fig:con_3d_res_slices}, we present more results of the 3D models across different layers. The first two columns show the two layers of the $4\times 4 \times 2$ 3D map which describes the target channel proportions for the $64\times64\times32$ images across the 3 dimensions. The remaining columns show generated 2D slices at different layers, namely the \nth{1}, \nth{8}, \nth{16} and \nth{24} layers. These results demonstrate that the 3D generated samples are spatially-correlated with the target maps in the 3 dimensions. We note here that although we show results as 2D slices, the 3D images are generated by a single-pass to the generator. 


\section{Quantifying Differences Between Generated Samples with the Same Conditioning Maps}
\label{sev_diff}
\begin{figure*}[h]
	\centering
	\includegraphics[width=0.9\linewidth]{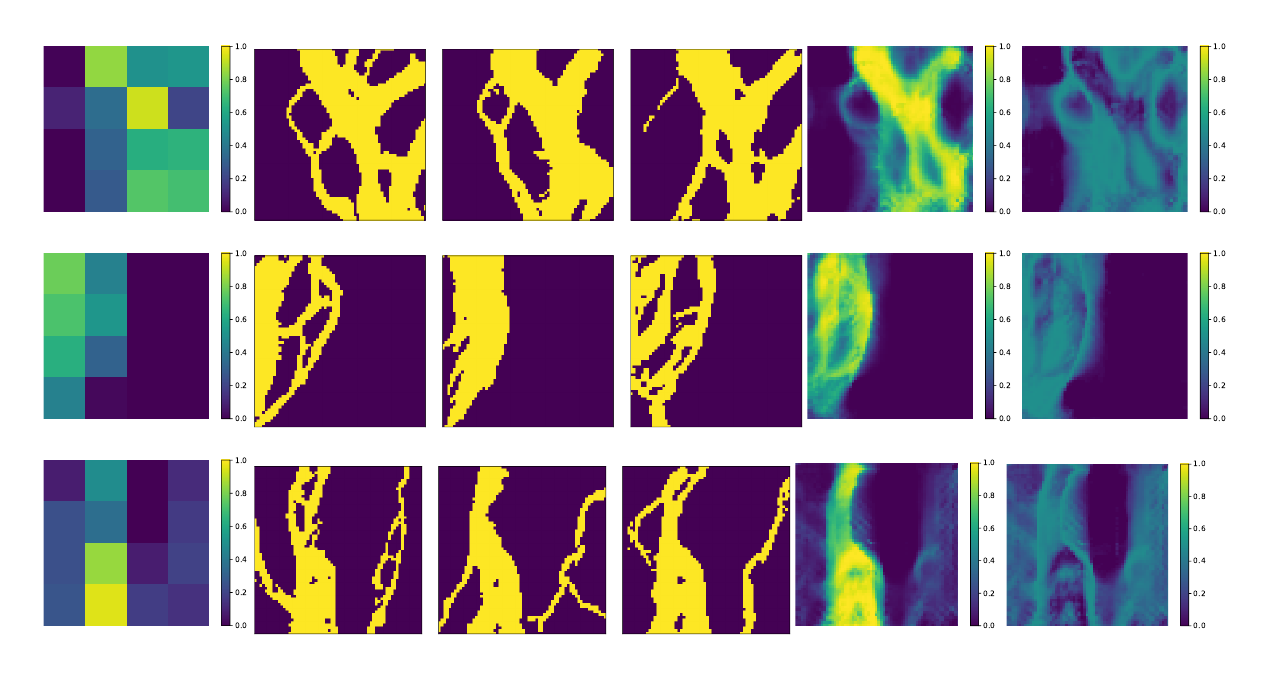}
	\caption{First column shows the target maps, the second, third and forth columns show generated non-stationary realizations using the target maps, the fifth columns shows the mean per pixel map and the standard deviation maps are shown in the rightmost column.}
	\label{fig:BR_std}
\end{figure*}

In this section, we present additional results for the braided river masks generated with the same condition maps but with different latent vectors and show the stochastic differences between the samples using standard deviation maps. As shown in Figure \ref{fig:BR_std}, the rightmost column shows standard deviation map which represents the stochastic variation around the mean value for each pixel. Small values mean low stochasticity (i.e., it is more likely for the pixel to be either a channel or background facies) . 

\section{Flow Simulation Results on Training and Generated Samples}
\label{sev_flow}
In this section, we present results of a flow simulation problem on the training samples showed on left side of Figure \ref{fig:datasets} and the corresponding generated samples showed on the top row in Figure \ref{fig:AR_res}. We consider the problem of a uniform flow where water is injected in order to displace contaminate in a subsurface reservoir. Flow is injected at the left side boundary and produced from the right side boundary and no-flow boundary conditions are imposed on the top and bottom sides. The problem formulation and settings are identical to those presented in \cite{chan2017parametrization}.

\begin{figure}
	\centering
	\includegraphics[width=0.9\linewidth]{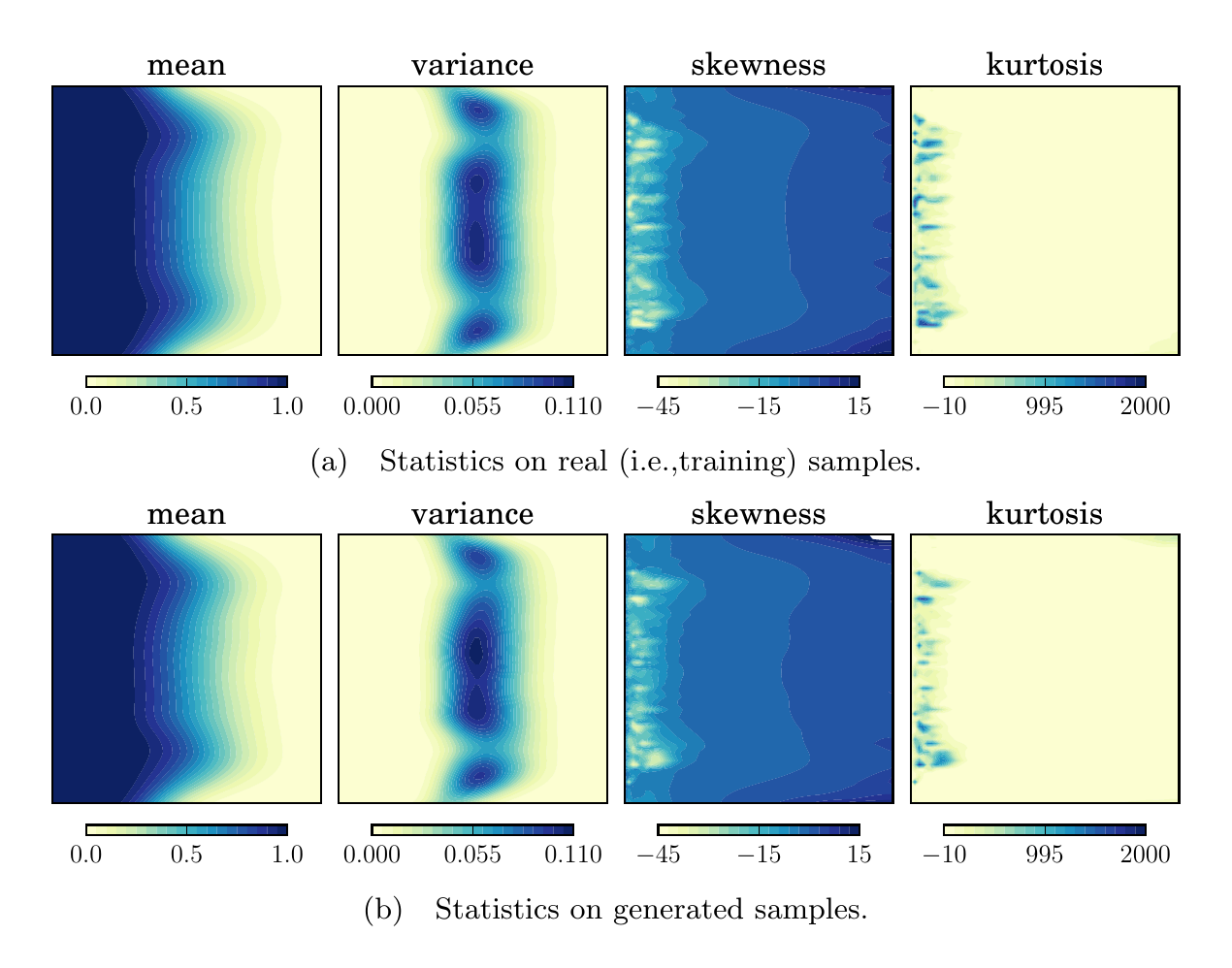}
	\caption{Saturation statistics of a uniform flow on training and generated samples at $t=0.5$ PVI.}
	\label{fig:flow_sat}
\end{figure}

\begin{figure}
	\centering
	\includegraphics[width=0.85\linewidth]{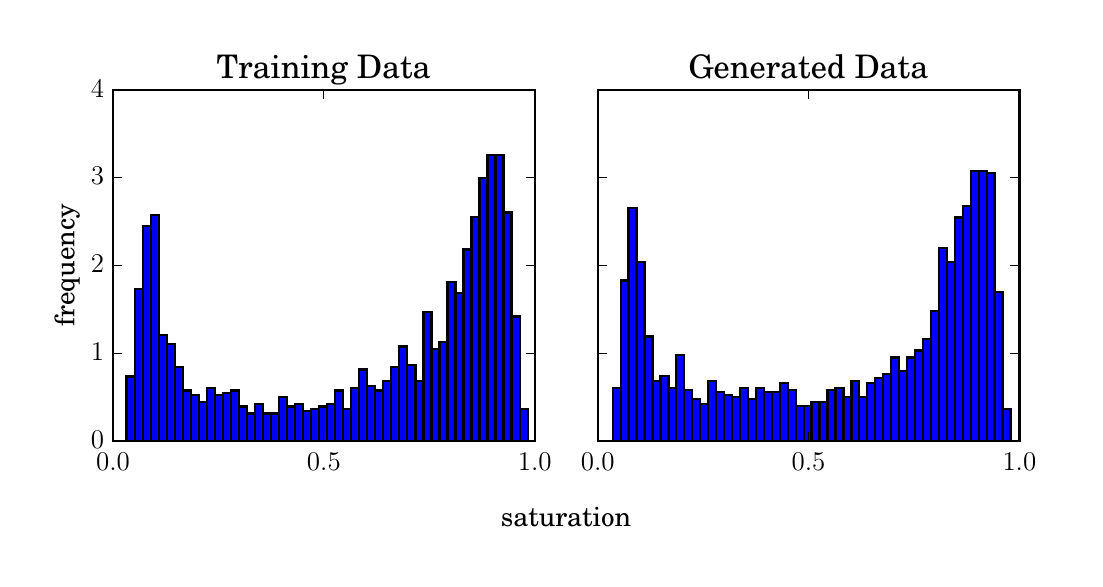}
	\caption{Saturation histogram at a fixed point in the domain where the saturation has the highest variance.}
	\label{fig:flow_sat_hist}
\end{figure}

We performed a total of 4000 flow simulations, 2000 corresponding to the training samples and 2000 simulations on the GANs generated samples. Flow statistics of the saturation map at $t=0.5$ PVI are shown in Figure \ref{fig:flow_sat} for the real and generated samples. As shown, statistics from generated realizations are very similar to the statistics from the training samples. Saturation histograms calculated at the point where the saturation has the highest variance are shown in Figure \ref{fig:flow_sat_hist}, where the two histograms from the training and generated samples match very well.

\begin{figure}
	\centering
	\includegraphics[width=0.9\linewidth]{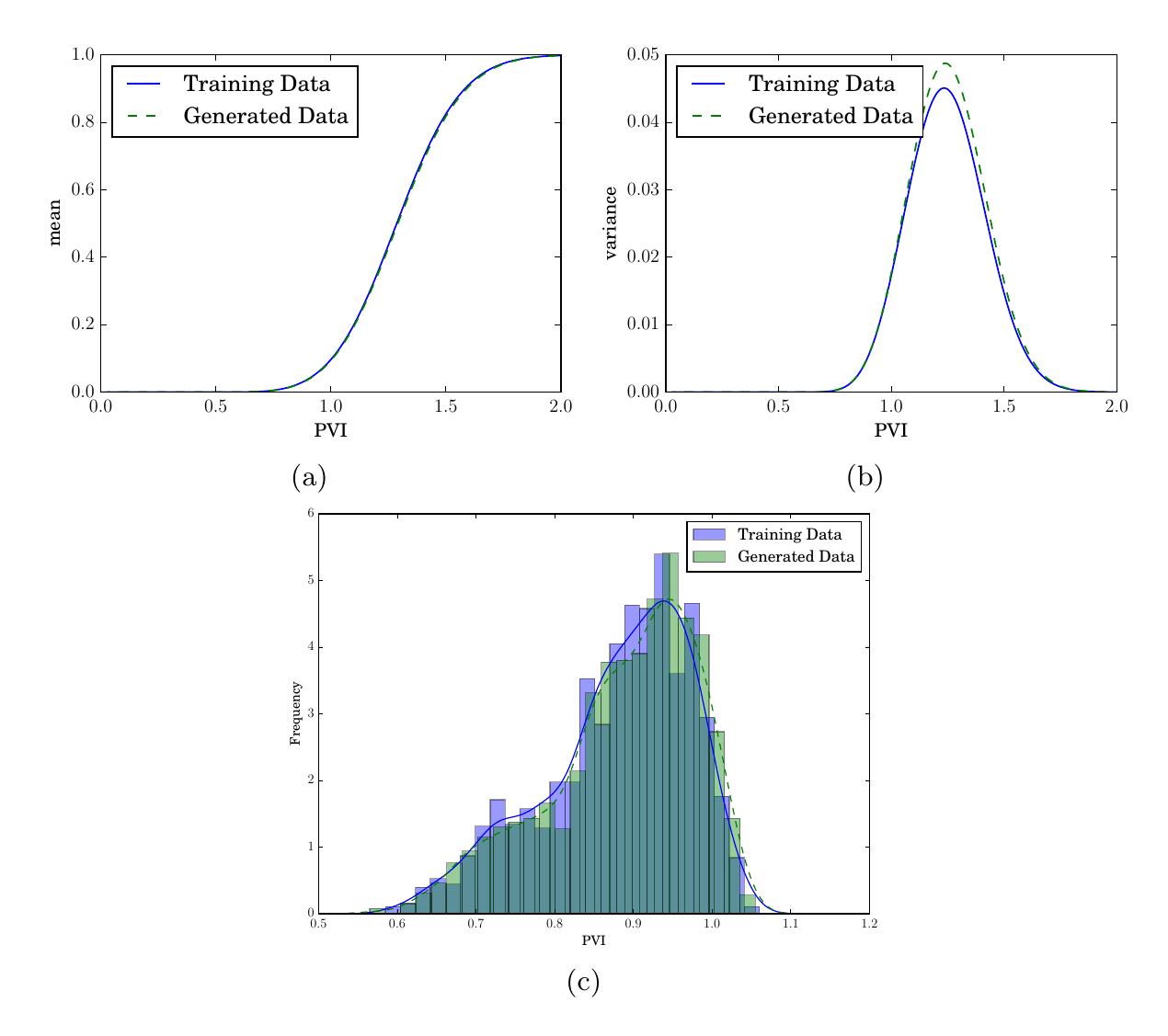}
	
	\caption{Production statistics of a uniform flow on training and generated samples at. (a) Mean of production curves (b) Variance of production curves (c) Histogram of water breakthrough times (expressed in pore volume injected).}
	\label{fig:production}
\end{figure}

Production curves statistics are shown in Figure \ref{fig:production}, where we calculated the mean and the variance of the production curves at different times. We have also plotted the histogram of the water breakthrough time (i.e., i.e., the time where the injected clean water level reaches the production well with a 1\% threshold). As we can see, the calculated statistics on the generated realizations showed very good agreement with the ones on the training samples which reflect the capabilities of the GANs models.

\end{document}